\documentclass{article}

\usepackage{microtype}
\usepackage{graphicx}
\usepackage{subfigure}
\usepackage{booktabs} % for professional tables

\usepackage[utf8]{inputenc}
\usepackage{cite}
\usepackage{amsmath,amssymb,amsfonts}
\usepackage{algorithmic}
\usepackage{graphicx}
\usepackage{textcomp}
\usepackage{xcolor}
\usepackage{natbib}
\usepackage[normalem]{ulem}
\usepackage{tabularx, multirow}
\usepackage{lipsum}
\usepackage{makecell}
\usepackage{url}
\usepackage{dirtytalk}
\usepackage{graphicx} % for horizontal figures

\usepackage{pdfpages}
\usepackage{hyperref}

\usepackage[accepted]{icml2023}

\usepackage{amsmath}
\usepackage{amssymb}
\usepackage{mathtools}
\usepackage{amsthm}

\usepackage[capitalize,noabbrev]{cleveref}

\theoremstyle{plain}

\theoremstyle{definition}

\theoremstyle{remark}

\usepackage[textsize=tiny]{todonotes}
\begin{document}

\twocolumn[
\icmltitle{THOS: A Benchmark Dataset for Targeted Hate and Offensive Speech}

% It is OKAY to include author information, even for blind
% submissions: the style file will automatically remove it for you
% unless you've provided the [accepted] option to the icml2023
% package.

% List of affiliations: The first argument should be a (short)
% identifier you will use later to specify author affiliations
% Academic affiliations should list Department, University, City, Region, Country
% Industry affiliations should list Company, City, Region, Country

% You can specify symbols, otherwise they are numbered in order.
% Ideally, you should not use this facility. Affiliations will be numbered
% in order of appearance and this is the preferred way.
\icmlsetsymbol{equal}{*}

\begin{icmlauthorlist}
\icmlauthor{Saad Almohaimeed}{UCF}
\icmlauthor{Saleh Almohaimeed}{UCF}
\icmlauthor{Ashfaq Ali Shafin}{FIU}
\icmlauthor{Bogdan Carbunar}{FIU}
\icmlauthor{Ladislau B\"{o}l\"{o}ni
}{UCF}
%\icmlauthor{Firstname6 Lastname6}{sch,yyy,comp}
%\icmlauthor{Firstname7 Lastname7}{comp}
%\icmlauthor{}{sch}
%\icmlauthor{Firstname8 Lastname8}{sch}
%\icmlauthor{Firstname8 Lastname8}{yyy,comp}
%\icmlauthor{}{sch}
%\icmlauthor{}{sch}
\end{icmlauthorlist}

\icmlaffiliation{UCF}{Department of Computer Science, University of Central Florida, Orlando, FL 32816, USA}

\icmlaffiliation{FIU}{Florida International University, Miami, FL 33199, USA}

\icmlcorrespondingauthor{Saad Almohaimeed}{almohaimeed.saad@gmail.com}
\icmlcorrespondingauthor{Ladislau B\"{o}l\"{o}ni}{ladislau.boloni@ucf.edu}

% You may provide any keywords that you
% find helpful for describing your paper; these are used to populate
% the "keywords" metadata in the PDF but will not be shown in the document
\icmlkeywords{Machine Learning, ICML}

\vskip 0.3in
]

% this must go after the closing bracket ] following \twocolumn[ ...

% This command actually creates the footnote in the first column
% listing the affiliations and the copyright notice.
% The command takes one argument, which is text to display at the start of the footnote.
% The \icmlEqualContribution command is standard text for equal contribution.
% Remove it (just {}) if you do not need this facility.

%\printAffiliationsAndNotice{}  % leave blank if no need to mention equal contribution

\printAffiliationsAndNotice %{\icmlEqualContribution} % otherwise use the standard text.

\let\cleardoublepage\clearpage
\begin{abstract}

Detecting harmful content on social media, such as Twitter, is made difficult by the fact that the seemingly simple yes/no classification conceals a significant amount of complexity.
Unfortunately, while several datasets have been collected for training classifiers in hate and offensive speech, there is a scarcity of datasets labeled with a finer granularity of target classes and specific targets. In this paper, we introduce THOS, a dataset of 8.3k tweets manually labeled with fine-grained annotations about the target of the message. We demonstrate that this dataset makes it feasible to train classifiers, based on Large Language Models, to perform classification at this level of granularity.

\textcolor{red} {\textit{\textbf{Warning:} Due to the nature of the research, the body of this paper contains offensive language. }}
\end{abstract}

\section{Introduction}
\label{Introduction}

Despite the obvious benefits of social media in connecting and entertaining people and providing a forum for public conversation about topics of interest to society, many social media platforms are plagued by harmful content (hate speech, cyberbullying, offensive statements). In recent years, it has been both a societal expectation, good business practice and sometimes a legal requirement for social media companies to filter harmful content. Due to the large number of messages on social media platforms, such a filtering can only be achieved through automatic classification. 

Early approaches to the task of identifying abusive content treated this as a classification into two or, perhaps, several classes, such as hate, offensive and neither \cite{davidson2017automated, waseem2016hateful}. Out of these, identifying offensive content using a dictionary-based approach is trivially simple. However, it has been recognized early on that the labels of offensive and hate speech need to be considered separately, as not every message containing swear words is hate speech, while a message without offensive words can be hateful and even incite violence. 

However, the seemingly straightforward classification of messages into two classes (harmful vs. not) obscures the fact that these are complex ideas that need to be identified and treated in the societal context. To mock somebody's favorite sports team is not hate speech; however, to mock somebody's religion, is. Making fun of somebody's sunburn  after a beach party is not hate speech, but in most contexts, joking about other people's skin color can be hateful. Finally, in many societies, the severity of an inappropriate statement depends on the target: for instance, negative statements directed at vulnerable minorities might be judged differently compared to similar messages targeting majority populations. 

Of course we could hope that a dataset of messages simply labeled as hate speech or not would allow machine learning models to infer features that implicitly take into account the target. However, it's unlikely that such an emergence will occur with small- or medium-sized datasets. Furthermore, we should also consider the fact that for many applications it is preferable to have a classifier that explicitly identifies the target of the message. For instance, the percentage of hate messages targeting a certain ethnic or religious group might be used to assess dangerous societal trends. 

To create the classifier algorithms for target detection, it is necessary to have well-annotated datasets at the appropriate level of resolution. While some datasets that include the target in the annotation exist~\cite{ousidhoum2019multilingual, mathew2021hatexplain}, they cover only some of the possible ways in which the target can be classified at a finer resolution. A logical next step would be to consider separately the {\em topic} of the message (such as religion, politics or country), and within a topic, a more specific {\em subtopic} such as China or India. Apart from the target of the message, the dataset would also be labeled as to whether it contains offensive words (OFF) and whether, in the human annotator's opinion, it contains hate speech (HS). 

The THOS dataset described in this paper aims to provide such a labeled collection of messages for training and validating fine-grained classifiers. In the remainder of this paper, we describe the structure of the dataset and the procedure used to create it. We also describe some initial results that use the dataset to train fine-grained classifiers based on large language models.

The main contributions of this paper are as follows:

\begin{itemize}

\item We created and manually labeled a Twitter dataset (THOS)\footnote{ https://github.com/mohaimeed/THOS \label{DatasetRepo}} for hate and offensive speech, which exhibits a finer level of granularity in its labeling compared to previous such datasets. For instance, THOS includes labels identifying the targets as general topics such as country, religion, ethnicity, politics or individuals, as well as specific subtopics such as the USA, China, Christian, Jewish, or Muslim.

\item We analyze the dataset with regards to the co-occurences of its labels. One of our findings confirms the conjecture that relying solely on offensive language detection, such as employing dictionary-based approaches, is insufficient in effectively tracking hate speech. Surprisingly, approximately 55\% of instances classified as hate speech do not contain offensive language, while a significant proportion of offensive language does not meet the criteria for hate speech.

\item We show that the THOS dataset, though of a moderate size, is sufficient to train fine-grain classifiers. We demonstrate that fine-tuning large language models (LLMs) on THOS can yield classifiers of sufficient accuracy for tracking trends of hate speech against specific subgroups on social media.

\end{itemize}
\section{Related Work}
\label{sec:RelatedWork}

As hate speech against specific groups emerged as a major problem on social media, a growing community of researchers started to focus on its detection and classification. These projects were motivated through various objectives, ranging filtering offensive language and the study of trends in hate against specific communities to complex problems such as the identification of hate campaigns organized by operatives of political organizations. 

The first generation of such systems used traditional machine learning techniques and focused on classifying the messages in a handful of classes. From the beginning, the creation and labeling of datasets were the primary prerequisites for the classification systems. \cite{waseem2016hateful} developed a dataset named NAACL\_SRW\_2016, consisting of 16k tweets, for multi-class classification focusing on sexism, racism, and normal speech. They explored the impact of various features, including the geographic location of the tweet and the gender of the user. The authors conducted experiments using logistic regression as the machine learning model and evaluated the model's effectiveness using F1-score, precision, and recall metrics. Their findings revealed that the best performance was achieved when utilizing character four-gram features combined with the gender feature.

\cite{davidson2017automated} have created a Twitter dataset (24.8k tweets) that considers the majority of the annotators' voting as multi-class labeling (i.e. hate, offensive, or neither). The authors use 1 to 3-gram features weighted by TF-IDF for the Part-of-Speech tags using Natural Language Toolkit (NLTK)
% \footnote{NLTK - \href{https://www.nltk.org/}{https://www.nltk.org/}}, 
along with the sentiment's score of each tweet using VADER~\cite{hutto2014vader}. They experimented with different machine learning models and with the best performance optained by logistic regression with L2 regularization. An interesting finding of the paper was that sexist content was mostly classified as offensive while homophobic and racist content was mostly detected as hate speech. 

\cite{waseem2016you} examined the impact of annotator knowledge by extending the NAACL\_SRW\_2016 dataset with 6.9k tweets labeled by four different annotators. The author found that the performance of the expert's annotation model was comparable to NAACL\_SRW\_2016, but including the labels of amateur annotators lowered the performance compared to the NAACL\_SRW\_2016 dataset on both binary and multi-class classification. 

\medskip

One of the focus of the second generation of projects was the recognition that the {\em target} of the speech also holds significant relevance in hate speech detection.

\cite{zampieri2019predicting} created an offensive language identification dataset consisting of 14.1k tweets that consider three levels of aspects (i.e. speech type, offense type, and target type). They created a baseline by experimenting with two types of  embedding, FastText\footnote{fastText - \href{https://github.com/facebookresearch/fastText}{https://github.com/facebookresearch/fastText}} and custom on a linear SVM using word unigram, pre-trained BiLSTM and CNN models by \cite{rasooli2018cross, kim-2014-convolutional}. 

\cite{ousidhoum2019multilingual} created a multi-lingual (Arabic, English, and French) hate speech dataset of 13k tweets that considers the aspects of directness, hostility, target, group and annotator. The dataset was used to train classifiers using both legacy approaches (Bag-of-Words with logistic regression) and deep learning based ones (BiLSTM using the \cite{ruder122017learning} learning network). The authors tested the model with different settings according to the number of tasks (number of labels to predict) and languages. They found that single-task single-language (STSL) model outperforms the multi-task multi-lingual (MTML) model in \textit{directness} aspect while MTML performs better than STSL in the other aspects. 

\cite{fortuna2020toxic} analyzed six public abusive language datasets by using FastText to investigate the compatibility between different datasets and their definitions of classes. The analysis found that there is a significant incompatibility between the definitions of classes by different datasets. 

\cite{mathew2021hatexplain} had extended previous datasets from \cite{davidson2017automated, ousidhoum2019multilingual} and \cite{mathew2019spread} to create the HateXplain dataset containing 20k Twitter and Gab annotated samples. The dataset considers hate and offensive as the two classes of abusive language, and also includes the target group. The dataset also contains an unusual label listing the tokens that represent the position of the abusive content in the text (i.e. {\em rationales}). The dataset was used to train CNN-GRU, BiRNN and BERT models and word embedding by GloVe for non-BERT models. The performance was evaluated through a variety of metrics such as accuracy, macro F1 and AUROC, the prediction bias (i.e. AUC) by \cite{borkan2019nuanced} and finally the explainability metrics (i.e. Plausibility and Faithfulness). It was found that the rationale labels can improve both performance and bias metrics, but not the explainability metric.

Table~\ref{tab:DatasetComparison} summarizes the size, granularity and annotator expertise of the datasets discussed above, positioning THOS in the context of this work. 

\begin{table*}[t]
\centering
\newcolumntype{P}[1]{>{\centering\arraybackslash}p{#1}}  
  \caption{Dataset Comparison (existing work)}
\begin{tabular}{ p{2.0cm} P{1.1cm}P{1.3cm} P{2.3cm} r P{1.0cm} r r r  c|c|c|c|c|c|c|c|c}
    \hline
    \thead{Dataset} & \multicolumn{2}{l}{\thead{Speech Type}} & \multicolumn{3}{l}{\thead{Target}} & \multicolumn{2}{l}{\thead{Annotator}} & \thead{Size} \\ 

    \thead{} & \thead{Hate} & \thead{ Offensive} & \thead{TPCs} & \thead{STPCs}& \thead{Indvs}& \thead{Expert} & \thead{Crowdsource\\ worker} & \thead{} \\
    \hline
    
    \cite{waseem2016hateful} &  & \checkmark &  &  &  & 16,907 &  & 16,907  \\

    \cite{waseem2016you} &  & \checkmark &  &  &  & unknown & 6,909  & 6,909\\

    \cite{davidson2017automated} & \checkmark & \checkmark &  &  &  &  & 24,783 & 24,783 \\

    \cite{zampieri2019predicting} &  & \checkmark & not explicit &  & \checkmark & 300 & 14,100 & 14,100 \\

    \cite{ousidhoum2019multilingual} & \checkmark &  \checkmark & religion, sex, gender, origin, disability & 15 &  & 300 & 13,000  & 13,000 \\

    \cite{mathew2021hatexplain} & \checkmark & \checkmark &  & 23 &  &  & 20,148 & 20,148  \\

    THOS (this paper) & \checkmark &  \checkmark & country, religion, ethnicity, politics & 31 & \checkmark & 8,282 &  &  8,282 \\
    \hline
  \end{tabular}
  \\[1em]
  \label{tab:DatasetComparison}
\end{table*}

\section{Dataset creation}
\label{sec:Dataset}

\subsection{Motivation and general principles}

The goal of the THOS dataset is to support research for the creation and evaluation of accurate and detailed classifiers of harmful speech on the internet. As Table~\ref{tab:DatasetComparison} shows, THOS seeks to address an existing gap in the collection of hate speech datasets by focusing on higher resolution labels with respects to the target and a moderate size, but higher quality labeling performed by experts. Like the majority of comparable datasets, THOS includes explicit labels of hate speech (HS) and offensive language (OFF). We will call the combination of HS and OFF as {\em explicit hate speech} (HSOFF), while the presence of the HS label without the OFF label as {\em implicit hate speech}. The dataset also includes labels identifying the target of the message at two levels of the resolution: the broader, more general topics (TPC) and the more specific sub-topics (STPC). To ensure the accuracy and reliability of the annotations, domain experts were engaged in the annotation process, rather than relying on crowd-sourcing. This approach ensures the uniformity of the annotations, resulting in a more robust and accurate dataset~\cite{waseem2016you}.

\subsection{Content of the dataset}
\label{sec:DatasetContent}

The THOS dataset contains 8,282 tweets, with 46 data fields, which include the text of the tweet, 5 meta-data fields relating to the identity of the tweet and annotator and 40 boolean labels. The labels can be grouped into four classes. 

The \textbf{Hate} label answers the questing whether the speech or text promotes or incites hatred toward an explicitly or implicitly targeted individual or group. The \textbf{Offensive} indicates whether the speech or text contains offensive (swear) words regardless whether it incites or promotes hate. 

The third class comprises six labels that encode TPCs. Based on previous research, which
identified ethnicity, origin(country), race and religion \cite{fortuna2018survey, relia2019race, scheitle2016religion} as the most common target groups in hate speech, these were added as available TPC selections. We decided to group the ethnicity/race in a single TPC. In the pilot phase of the dataset creation, we also identified as large classes of hate targets {\em political groups} and {\em individuals}. Finally, a TPC representing {\em other} was added to capture targets outside these groups. These labels are not intended to be mutually exclusive: for instance, 1166 tweets in THOS fall under two TPCs, while 66 under three TPCs. 

The fourth class of labels in the THOS dataset are the 31 labels describing STPCs. As shown in Table~\ref{tab:STPCsDistribution}, STPCs can be viewed as a second level of classification hierarchy after TPCs. However, there is also a qualitative difference: while TPCs delineate targets in terms of high level {\em concepts}, most STPCs correspond to {\em concrete} instantiations of those concepts, such as specific countries, ethnicities or political orientations. 

Table~\ref{tab:DatasetSamples} shows several examples from the dataset and the associated labels.

\subsection{Data acquisition and labeling}

\begin{table}
\centering
\newcolumntype{P}[1]{>{\centering\arraybackslash}p{#1}}
\caption{Speech Type Distribution}
\begin{tabular}{ l r r } 
\hline
\thead{Speech Type} & \thead{\# of samples} & \thead{Percentage \%}\\ 
\hline
Hate only & 1878 & 22.7   \\ 
Offensive only & 933 & 11.3   \\ 
Hate and Offensive & 1523 & 18.4   \\ 
Normal & 3948 & 47.6   \\ 
\hline
\end{tabular}
\label{tab:SpeechTypeDistribution}
\end{table}

\begin{table}
\centering
  \caption{TPCs Distribution}
\newcolumntype{P}[1]{>{\centering\arraybackslash}p{#1}}
\begin{tabular}{ l r r r }
\hline
\thead{TPC} & \thead{\# of shared \\  samples} & \thead{\# of samples} & \thead{total}\\ 
\hline
Country & 914 & 544 & 1458   \\ 
Religion & 384 & 394 & 778   \\ 
Ethnicity/Race & 1024 & 810 & 1834   \\ 
Politics & 125 & 42 & 167   \\ 
Other & 1303 & 1771 & 3074  \\ 
Individuals & 1388 & 2281 & 3669   \\ 
\hline
Total & 5138 & 5842 & - \\
\hline
\end{tabular}
\label{tab:TPCs Distribution}
\end{table}

The tweets forming the THOS dataset were collected through the Twitter API using the Tweepy\footnote{Tweepy - \href{https://www.tweepy.org/}{https://www.tweepy.org/}} library. To avoid collecting tweets that don't fit under any of the TPCs, we used keyword search to look up tweets that are relevant to \textbf{countries} \textit{(i.e. USA, UK, China, and India)}, \textbf{religions} \textit{(i.e. Muslim, Christian, and Jewish)} as well as \textbf{ethnicities} and races \textit{(i.e. Black, White, Asian, Arab, European, African, Hispanic, and Latino)}. The keywords were combined with prefixes such as \textit{all, all of the, most, most of the some, some of the}. We collected a total of 31k tweets using the process.

To align the THOS dataset with the existing datasets, we started by creating the definitions for hate speech and offensive speech drawing on the ones used by other datasets in the literature. Starting with these definitions, we proceeded with a pilot phase, during which 280 sample tweets have been carefully annotated. These annotated samples were then provided to the expert annotators for study, revision, and feedback. Using this feedback we created a rules and process document (RPD)\footref{DatasetRepo} to serve as a guideline for the experts to resolve potential ambiguities in the content.

% HERE LOTZI

The annotation of the dataset was carried out by five experts. Each expert was assigned a 3k tweets to annotate, out of which they were required to submit 1.5k tweets. For THOS, the quality of the annotation was prioritized over coverage, thus experts were allowed to skip tweets that are unclear or meaningless, to enhance the overall accuracy of the annotations. The annotators reported an average annotation rate of 80 tweets/hour (45s/tweet). After every 200 submitted tweets, the annotations were reviewed and checked for conformance to the RPD. 

As a note, the classification of hate speech did not check whether the tweet refers to a fake news or not, as the fact-checking of the tweet would be highly expensive.

\begin{table}
\centering
  \caption{STPCs Distribution}
\newcolumntype{P}[1]{>{\centering\arraybackslash}p{#1}}
\begin{tabular}
{p{0.2cm} p{2.0cm} r r r  } 
\hline
     & \thead{STPC} & \thead{\# of shared\\samples} & \thead{\# of\\samples} & \thead{total}\\ 
\hline
\parbox[t]{2mm}{\multirow{6}{*}{\rotatebox[origin=c]{90}{Country}}} &
 USA & 362 & 186 & 548\\
 & UK & 66 & 34 & 100\\
 & China & 232 & 80 & 312\\
 & India & 113 & 48 & 161\\
 & Other\_Country & 401 & 115 & 516 \\
 & Undefined\_Country & 4 & 2 & 6 \\
 \hline
 \parbox[t]{2mm}{\multirow{5}{*}{\rotatebox[origin=c]{90}{Religion}}}
 & Muslim & 126 & 105 & 231\\
 & Christian & 126 & 169 & 295\\
 & Jewish & 154 & 110 & 264\\
 & Other\_Religion & 14 & - & 14 \\
 & Undefined\_Religion & 1 & - & 1 \\
\hline
\parbox[t]{2mm}{\multirow{10}{*}{\rotatebox[origin=c]{90}{Ethnicity/Race}}}
 & White & 314 & 124 & 438\\
 & Black & 240 & 78 & 318\\
 & Arab & 127 & 94 & 221\\
 & African & 88 & 54 & 142\\
 & European & 101 & 73 & 174\\
 & Asian & 203 & 104 & 307\\
% & Latino & 97 & 54 & 151\\
% & Hispanic & 58 & 32 & 90\\
 & Hispanic & 155 & 86 & 241 \\
 & Other\_Ethnicity & 163 & 81 & 244 \\
 & Undefined\_Ethnicity & - & 1 & 1 \\
\hline
\parbox[t]{2mm}{\multirow{4}{*}
{\rotatebox[origin=c]{90}{Politics}}}
 & Republican & 37 & 18 & 55\\
 & Democrat & 35 & 9 & 44\\
 & Other\_Politics & 66 & 12 & 78 \\
 & Undefined\_Politics & - & - & - \\
\hline
\parbox[t]{2mm}{\multirow{5}{*}{\rotatebox[origin=c]{90}{Other}}}
 & People & 24 & 66 & 90\\
 & Male & 168 & 252 & 420\\
 & Female & 288 & 353 & 641\\
 & Other\_Other & 1008 & 828 & 1836 \\
 & Undefined\_Other & 97 & 164 & 261 \\
 \hline
 \parbox[t]{2mm}{\multirow{2}{*}{\rotatebox[origin=c]{90}{Indv}}}
 & Someone & 1148 & 1889 & 3037 \\
 & Undefined\_Individual & 385 & 329 & 714 \\
 \hline
& Total & 6246 & 5464 & - \\
\hline
\end{tabular}
\label{tab:STPCsDistribution}
\end{table}

\subsection{Comments on the class distribution in the dataset}

As we discussed before, the THOS dataset was designed as a tool for training and validating hate speech classifiers. The tweets we annotated were extracted using keywords that are known to refer to targets frequently subjected to hate speech. The objective was to create a rough balance between tweets with harmful content and those without. As Table~\ref{tab:SpeechTypeDistribution} shows, about half of the messages are marked as hate, offensive or both. We need to emphasize that this is not intended as an accurate representation of the messages in Twitter.

Likewise, we sought to achieve a roughly equal distribution of the TPCs (see Table~\ref{tab:TPCs Distribution}) and STPCs (see Table~\ref{tab:STPCsDistribution}). However, the distribution over these classes proved to be more uneven. At the TPC level, for instance, we have a significantly smaller number of tweets in the Politics TPC, as the method for searching for tweets did not explicitly search for this topic. For the STPCs, a substantial number of tweets were classified under the labels prefixed with "Other", denoting values specified but outside the list of the enumerated ones, and the "Undefined", where the topic implied the existence of a particular STPC that, however, was not defined in the tweet. 

Lastly, we did not label STPCs with specific individuals. Thu, here we have only two classes - when an individual was named (the STPC Somebody) and when an individual is implied but not identifiable from the tweet (the STPC Undefined Individual).

%---------- Correlation Table
\begin{table}
\centering
\caption{HSOFF and TPCs correlation}
\newcolumntype{P}[1]{>{\centering\arraybackslash}p{#1}}
\begin{center}
\begin{tabular}{ p{1.5cm} r r r r } 
\hline
\thead{TPCs} &  \thead{Normal}&\thead{HS} & \thead{OFF} & \thead{HSOFF}\\ 
\hline
Country & 615 & 426 & 138 & 279 \\ 
Religion & 314 & 211 & 71 &  182  \\ 
Ethnicity & 750 & 394 & 252 & 438  \\ 
Politics   & 55 & 82 & 0 &  30  \\ 
Individual  & 1927 & 901 & 334 & 507 \\
\hline
\end{tabular}
\\[1em]
\end{center}
\label{tab:HSOFFandTPCsCorrelation}

\end{table}

\begin{table*}
\centering
  \caption{Dataset samples}
\newcolumntype{P}[1]{>{\centering\arraybackslash}p{#1}}

\begin{tabular}{p{9.8cm}  P{1.8cm} P{1.7cm} P{0.7cm} P{0.7cm} c|c|c|c|c|c|}
    \hline
    \thead{Tweet} & \thead{TPCs}& \thead{STPCs} & \thead{HS} & \thead{OFF} \\
    \hline
    @USER It is Allah's anger, we need to repent and make tawba all muslims of the world. & Religion & Muslim & NO & NO \\
    \hline
    @USER she Hispanic? cuz my mom would say some sh*t like that lol & Ethnicity, Individual & Hispanic, Undefined & NO & YES \\
    \hline
    African immigrants all have the same annoying accent lol
     & Ethnicity, Other & African, Other\_Other & YES & NO \\
    \hline
    @USER They always lie & Other & Undefined & YES & NO \\
    \hline
    @USER It is a question that all Christians must always ask themselves. & Religion & Christian & NO & NO \\
    \hline
    @USER Latinos can’t have sh*t. & Ethnicity & Hispanic & NO & YES \\
    \hline
    Why the f*ck do I have an Asian teacher for a English teacher tf is this Tom foolery. & Ethnicity, Other & Asian, Teacher & YES & YES \\
    \hline
    @USER He is one of the China a*s kissers in DC. & Country, Individual & China, Undefined & YES & YES \\
    \hline
    
  \end{tabular}
  \label{tab:DatasetSamples}

\end{table*}

\section{Experimental Study}
\label{sec:ExperimentalStudy}

In this section, we verify that the THOS dataset is suitable for its intended purpose, namely, serving as training and validation data. We anticipate that the use of THOS will improve  the detection of various types of harmful content on social media in at least two ways. First, it should enable fine-grain classification of the various types of hateful and offensive speech, identifying the various TPCs and STPCs. This is important for the social media companies and society in general, because different targets might represent different levels of risk, and require different societal responses. Second, the fine-grained annotations of the training data might help classifiers develop the appropriate inductive bias that would allow them to achieve a more accurate classification, even if the desired output is only a two-way or four-way classification along the hate and offensiveness axes. 

The development of such new generations of classifiers are beyond the scope of this paper. However, it is incumbent upon us that, to validate the value of the dataset, to verify whether it can serve as appropriate training and validation data for such classifiers. 

\subsection{Experimental setup}

The current state-of-the-art in text classification relies on the use of large language models (LLMs). One frequently used approach is based on starting from a foundation model and fine-tuning it to predict a specific label through a linear layer. LLMs such as BERT~\cite{devlin2018bert}, RoBERTa~\cite{liu2019roberta} and the Twitter-specialized BERTweet~\cite{nguyen2020bertweet} all offer the possibility to train classifiers following this general idea with relatively modest computational expenditure. As a note, for the newest generation of very large LLMs, such a training model would not be feasible. Therefore, different training approaches would be necessary, the discussion of which is beyond the scope of this paper. 

In order to perform experiments with the dataset we implemented multi-label classifiers based on BERT, RoBERTa and BERTweet. These models were fine-tuned over the course of 20 epochs, with a batch size of 16, the learning rate of 2e-5 and maximum sequence length of 200.

Regarding the metrics used to evaluate the classifiers, we note that while the dataset is well-balanced with regards to the hate (HS) and offensive (OFF) labels, it is unbalanced in terms of the TPCs and even less so with regards to the individual STPCs. Thus, in our experiments, in addition to the cross-entropy loss, we also reported the macro-averaged F1 and AUROC scores, which are more fair in an unbalanced dataset. These values had been independently calculated for each TPC and STPC and then averaged over all the TPCs and STPCs, respectively. 

\subsection{Experimental results}

% LOTZI: Here

The experiments we conducted aimed to predict the hate and offensive (HSOFF), TPC, and STPC values using the three models we trained. We considered three separate scenarios. 

In the first scenario, the input of the neural network was the text of the tweet, while the outputs were the HSOFF, TPC, and STPC values. This scenario is typically encountered when the social media network company or external observers are evaluating the messages based on only the text. The results are shown on the top row of Table~\ref{tab:ModelsExperimentPerformance}. We can make several observations about the results. First, the quality of the classification is significantly above random, validating the fact that the THOS dataset can be used to train classifiers, both for HSOFF as well as the finer grain TPC and STPC values. In fact, we found that the TPCs classification has a higher accuracy than the HSOFF values. This is likely due to the fact that the definitions of hate and offensive are more complex and open to interpretation. Overall, the accuracy values might not be sufficient for applications such as filtering (except maybe as a system that tags possible harmful messages to be verified by a human), but they could be instrumental for the study of hate speech trends. 

In the second scenario, we considered the case where the classifier receives both the text and the TPCs as input. From a practical point of view, this corresponds to instances where we have access to an oracle that can provide the TPC. In practice, such a prediction might come from the context - for instance, in social networks such as Reddit, the subreddit in which a message was posted can often determine the TPC with high accuracy. The results of this experiment are shown in the middle row of Table~\ref{tab:ModelsExperimentPerformance}. As expected, supplying the TPC as input boosts the accuracy of both the HSOFF classification (marginally), as well as the STPC classification (significantly). 

Lastly, in the third scenario, we investigated the case when the classifier receives the text and the STPCs as input (Table~\ref{tab:ModelsExperimentPerformance}, bottom row). We observe that this scenario also marginally increases the accuracy of the HSOFF classification. As expected, in this case, the prediction of the TPC is almost perfect, as the settings of the STPC essentially predicts that the encompassing TPC will also be set.

\begin{table*}[t]
\centering
  \caption{Experimental results: classification of HSOFF, TPCs, STPCs based on various inputs}
  \newcolumntype{P}[1]{>{\raggedleft\arraybackslash}p{#1}}
\begin{tabular}{ p{2.2cm} p{1.5cm}  r r r p{0.1cm} r r r p{0.1cm} r r r c|c|c|c|c|c|c|c|c|c|c|c|c|c }
    \hline
        \multicolumn{2}{l}{\thead{}} & \multicolumn{3}{l}{\thead{HSOFF}} & \thead{} & \multicolumn{3}{l} {\thead{TPCs}} & \thead{} & \multicolumn{3}{l}{\thead{STPCs}}\\

    \thead{Model} & Input &  \thead{F1}& \thead{AUROC} &  \thead{CE}  & &  \thead{F1} &  \thead{AUROC} &  \thead{CE} &  & \thead{F1}&  \thead{AUROC} &  \thead{CE} \\
    \hline
    BERT & txt & 0.78 & 0.81 & 1.23 & & 0.87  & 0.92 & 0.27 & & 0.76  & 0.88 & 0.07\\ 
    RoBERTa & txt & 0.77 & 0.81 & 1.20 & & \textbf{0.89} & \textbf{0.94} & 0.26 & & 0.70 & 0.85 & 0.07\\ 
    BERTweet & txt & 0.78 & 0.82 & \textbf{1.06} & & 0.88 & 0.93 & \textbf{ 0.23} & & 0.64 & 0.82 & 0.07\\ 
    \hline

    BERT & txt+TPCs & 0.78 & 0.81 & 1.23 & & - & - & - &  &\textbf{0.86} & \textbf{0.93} & \textbf{0.04} \\ 
    RoBERTa & txt+TPCs & \textbf{0.79} & \textbf{0.83} & 1.11 & & - & - & - & & 0.82 & 0.90 & \textbf{0.04} \\ 
    BERTweet & txt+TPCs & 0.78 & 0.82 & 1.08 & & - & - & - & & 0.70 & 0.84 & 0.05 \\ 
    \hline

    BERT & txt+STPCs & 0.78 & 0.82 & 1.19 & & 0.998 & 0.999 & 0.004 & & - & - & -  \\ 
    RoBERTa & txt+STPCs & \textbf{0.79} & 0.82 & 1.13 & & 0.998 & 0.999 & 0.002 & & - & - & -  \\ 
    BERTweet & txt+STPCs & 0.78 & 0.81 & 1.10 & & 0.999 & 0.999 & 0.001 & &  - & - & - \\ 
    \hline

  \end{tabular}
  \\[1em]
\label{tab:ModelsExperimentPerformance}
\end{table*}

For the second experiment, we know that for every STPC there is only one parent TPC. However, we believe that the confusion between the countries and ethnicities (e.g. European, African, Arab, Asian), is the reason behind not getting a 100\% score in TPCs classification since these kinds of STPCs depend on another token in the tweet text. Similar to the first experiment in the correlation experiment, the STPCs improved the HSOFF F1\_Macro score by 2 points while improving the AUROC\_Macro by 1 point to be 0.82.
\section{Conclusion}
\label{sec:Conclusion}

In this paper, we have proposed THOS, a new benchmark hate speech dataset consisting of 8.3k tweets. The dataset considers the hate and offensive speech along with the speech's target at two levels of resolution. In an experimental study, we demonstrated that the dataset can be successfully used to train LLM-based classifiers for the hate/offensive classification as well as the target classification. Furthermore, we found that knowing the target labels can improve the classification accuracy along the hate/offensive dimensions.

{\bf Acknowledgements:} This work was partially supported by the National Science Foundation under awards 2114948 and 2114911. 

% In the unusual situation where you want a paper to appear in the
% references without citing it in the main text, use \nocite
%\nocite{langley00}

%\bibliography{icml2023/References}
\bibliography{References}
\bibliographystyle{icml2023}

%%%%%%%%%%%%%%%%%%%%%%%%%%%%%%%%%%%%%%%%%%%%%%%%%%%%%%%%%%%%%%%%%%%%%%%%%%%%%%%
%%%%%%%%%%%%%%%%%%%%%%%%%%%%%%%%%%%%%%%%%%%%%%%%%%%%%%%%%%%%%%%%%%%%%%%%%%%%%%%
% APPENDIX
%%%%%%%%%%%%%%%%%%%%%%%%%%%%%%%%%%%%%%%%%%%%%%%%%%%%%%%%%%%%%%%%%%%%%%%%%%%%%%%
%%%%%%%%%%%%%%%%%%%%%%%%%%%%%%%%%%%%%%%%%%%%%%%%%%%%%%%%%%%%%%%%%%%%%%%%%%%%%%%

%\newpage
%\appendix
%\onecolumn

%\section{Appendix}

%\includepdf[pages={-}]{PRD.pdf}

%%%%%%%%%%%%%%%%%%%%%%%%%%%%%%%%%%%%%%%%%%%%%%%%%%%%%%%%%%%%%%%%%%%%%%%%%%%%%%%
%%%%%%%%%%%%%%%%%%%%%%%%%%%%%%%%%%%%%%%%%%%%%%%%%%%%%%%%%%%%%%%%%%%%%%%%%%%%%%%

\end{document}